\documentclass[10pt,twocolumn,letterpaper]{article}

\usepackage{wacv}              % To produce the CAMERA-READY version
\usepackage{graphicx}
\usepackage{amsmath}
\usepackage{amssymb}
\usepackage{booktabs}
\usepackage{array}
\usepackage{multirow}
\usepackage{makecell}
\usepackage{bbm}
\usepackage{color, colortbl}
\usepackage{arydshln}
\usepackage{xcolor}
\definecolor{Gray}{gray}{0.9}
\def\hm{\textcolor{black}}
\usepackage[accsupp]{axessibility}
\usepackage[pagebackref,breaklinks,colorlinks]{hyperref}

\usepackage[capitalize]{cleveref}
\crefname{section}{Sec.}{Secs.}
\Crefname{section}{Section}{Sections}
\Crefname{table}{Table}{Tables}
\crefname{table}{Tab.}{Tabs.}

\def\wacvPaperID{238} % *** Enter the WACV Paper ID here
\def\confName{WACV}
\def\confYear{2025}

\begin{document}

%%%%%%%%% TITLE - PLEASE UPDATE
\title{A Simple-but-effective Baseline for Training-free Class-Agnostic Counting}

\author{Yuhao Lin\footnotemark[1] , Haiming Xu\thanks{Equally contributed.} , Lingqiao Liu, Javen Qinfeng Shi\\
Australian Institution for Machine Learning\\
The University of Adelaide\\
{\tt\small yuhao.lin01,hai-ming.xu,lingqiao.liu,javen.shi@adelaide.edu.au}
% For a paper whose authors are all at the same institution,
% omit the following lines up until the closing ``}''.
% Additional authors and addresses can be added with ``\and'',
% just like the second author.
% To save space, use either the email address or home page, not both
% \and
% Second Author\\
% Australian Institution for Machine Learning\\
% The University of Adelaide\\
% {\tt\small secondauthor@i2.org}
}
\maketitle
% \footnotetext[1]{Yuhao Lin and Haiming Xu contributed equally to this work.}

%%%%%%%%% ABSTRACT
\begin{abstract}
  Class-Agnostic Counting (CAC) seeks to accurately count objects in a given image with only a few reference examples. While previous methods achieving this relied on additional training, recent efforts have shown that it's possible to accomplish this without training by utilizing pre-existing foundation models, particularly the Segment Anything Model (SAM), for counting via instance-level segmentation. Although promising, current training-free methods still lag behind their training-based counterparts in terms of performance.
In this research, we present a straightforward training-free solution that effectively bridges this performance gap, serving as a strong baseline. The primary contribution of our work lies in the discovery of four key technologies that can enhance performance. Specifically, we suggest employing a superpixel algorithm to generate more precise initial point prompts, utilizing an image encoder with richer semantic knowledge to replace the SAM encoder for representing candidate objects, and adopting a multiscale mechanism and a transductive prototype scheme to update the representation of reference examples. By combining these four technologies, our approach achieves significant improvements over existing training-free methods and delivers performance on par with training-based ones.

\end{abstract}

%%%%%%%%% BODY TEXT
\section{Introduction}
\label{sec:intro}
Class-Agnostic Counting (CAC)~\cite{lu2019class} is increasingly gaining recognition in the field of computer vision. In contrast to traditional class-specific object counting~\cite{ranjan2021learning,shi2022represent}, which relies on training models for specific predefined categories, CAC adopts a more flexible approach. It expects the model to adapt to a diverse array of objects across various classes using only a minimal number of examples~\cite{lu2019class}. This adaptability makes CAC a compelling area of research, offering potential for broad application in scenarios where rapid or dynamic object recognition is required.

To attain the ability for Class-Agnostic Counting, a common approach is to utilize a training dataset containing various objects along with their corresponding density maps. This training data is used to train a model that compares visual features between reference examples and the query image~\cite{lu2019class,shi2022represent,Dukic_2023_ICCV}. This methodology enables class-independent adaptation, which is essential for scenarios requiring dynamic counting. However, a significant drawback of these methods is their reliance on large-scale annotated training data, which presents challenges in terms of labor intensity and scalability across diverse visual categories~\cite{ma2023can}.
\begin{figure}
\centering
\includegraphics[width=\linewidth]{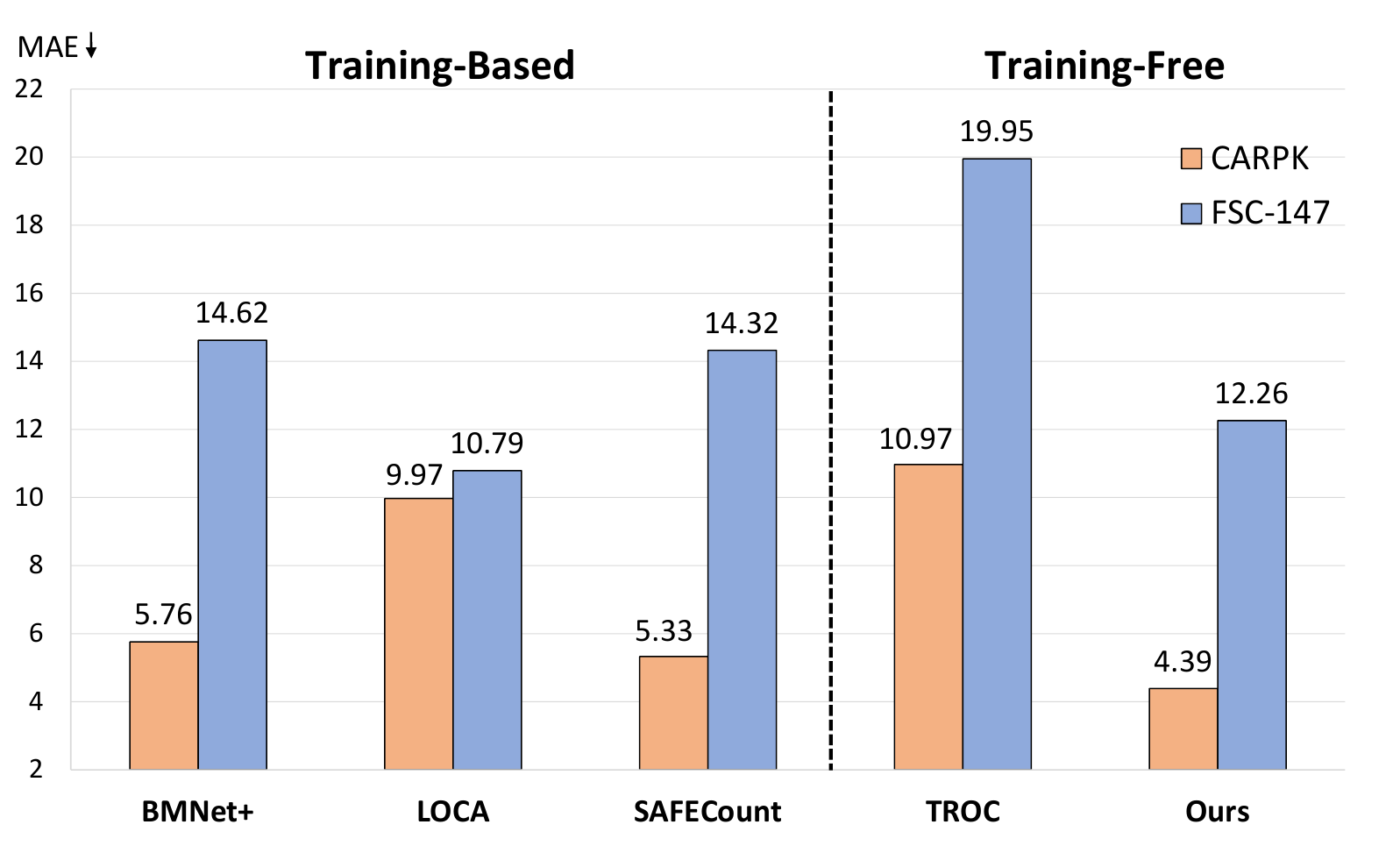}  
\caption{
% Performance visualization of our approach and both training-based and training-free approaches. It shows that our approach significantly reduces the counting performance gap between training-free methods and the training-based counterpart. Our approach even outperforms the best training-based methods on the CARPK dataset which demonstrates the potential of training-free methods.
Visualizing the performance, our method notably narrows the gap in counting accuracy between training-free and training-based approaches. Impressively, it even surpasses the top-performing training-based methods on the CARPK dataset, underscoring the potential of training-free approaches.
}
\label{fig:first}
\vspace{-0.5cm}
\end{figure}
Recently, the Segment Anything Model (SAM)~\cite{kirillov2023segany} has garnered substantial attention across various research domains due to its remarkable ability to generate high-quality object masks for any given image. This versatility has led to its application in numerous research areas, as evidenced by multiple studies ~\cite{yang2023track,ma2023segment,wang2023caption,shen2023anything}.
In the context of CAC, researchers have started adopting SAM for producing instance-level segmentation masks corresponding to referenced objects. By counting the number of instance masks, they can determine the final count of objects. 

Thanks to SAM's robust instance segmentation capabilities, this counting-via-instance-segmentation approach can function effectively even when utilizing the off-the-shelf SAM model without any additional training. Training-free methods hold appeal for CAC since they eliminate the need for a dedicated training dataset and alleviate the challenges associated with collecting extra training data. Nevertheless, despite their promise, current training-free CAC methods still lag behind their training-based counterparts in terms of performance, e.g., as demonstrated in Figure~\ref{fig:first}, 
% the existing best training-free method is 84.89\% and 105.82\% worse than the best training-based ones on FSC-147 and CARPK datasets respectively.
the current top training-free method falls short of the best training-based approaches by 84.89\% and 105.82\% on the FSC-147~\cite{ranjan2021learning} and CARPK~\cite{hsieh2017drone} datasets, respectively.

% the current top training-free method falls short of the best training-based approaches by 84.89\% and 105.82\%\footnote{\color{blue}{$\frac{10.97-5.33}{5.33}=1.058$, it indicates that the errors more than doubled.}} on the FSC-147~\cite{ranjan2021learning} and CARPK~\cite{hsieh2017drone} datasets, respectively.

In our research, we introduce an improved approach to achieve training-free Class-Agnostic Counting using SAM. Our method provides a straightforward yet robust means to significantly narrow the performance gap often observed in training-free CAC methods. This positions our approach as a valuable benchmark for future research in this domain.
Our approach to CAC relies on four pivotal technological components, each playing a crucial role:
First, we emphasize the paramount importance of precise point prompt placement within SAM to achieve accurate object masking. To this end, we advocate the utilization of superpixels, which enables the creation of initial point prompts, resulting in high object recall without imposing significant computational overhead.
Moreover, our thorough investigations have exposed limitations in SAM's current image encoder concerning its ability to effectively capture semantic-level similarity. In response, we propose the adoption of image encoders equipped to capture richer semantic information. This proposition holds the potential to enhance the representation of objects during the counting process.
Additionally, we recognize the significance of implementing a multiscale mechanism, particularly for the precise counting of small objects. By incorporating this mechanism, our approach accommodates variations in object size across diverse instances, thus yielding more accurate counting results.
Finally, we introduce a transductive prototype update strategy for reference examples. Through the augmentation of reference examples with candidate objects that are likely to share the same object category, we substantially enhance the reference-candidate similarity metric. This enhancement empowers us to more effectively determine whether an object proposal qualifies as the target of interest, ultimately contributing to the overall accuracy of our Class-Agnostic Counting methodology. Collectively, these four technological advancements serve as the cornerstone of our approach.

% Through rigorous experimentals, we have demonstrated the remarkable performance of our proposed methods and the effectiveness of the four key components. Significantly, we have demonstrated for the first time that a Class-Agnostic Counting (CAC) model, without training, can match the performance of traditional, training-based CAC approaches.

Our rigorous experiments demonstrated the remarkable performance of our proposed methods and the effectiveness of the four key components. Significantly, we show for the first time that a Class-Agnostic Counting (CAC) model without any training can achieve performance on par with that of training-based CAC approaches.

%-------------------------------------------------------------------------
\section{Related Work}
\label{sec:Related_Work}

\subsection{Class-Agnostic Counting Methods}
% class-specific object counting -> class-agnostic counting (learning based methods + training free methods)
% say some drawbacks: costly annotations, performance gap to learning-based methods
% The concept of class-agnostic counting is particularly significant for its flexibility, allowing for the counting of arbitrary object classes in images using support exemplars, as opposed to class-specific counting which relies on recognition of predefined objects in training datasets. This paradigm shift, referenced across various studies [5, 13, 16, 17, 19, 20, 24, 27, 28, 30–32], fosters a more dynamic application of counting models, enabling customization and adaptation to user-defined objects of interest.

Lu et al.~\cite{lu2019class} first address CAC, employed convolutional neural networks to extract and concatenate features from both query images and exemplars to estimate object counts. However, this method faced challenges related to overfitting when directly regressing from concatenated features. Addressing this, CFOCNet~\cite{you2023few} introduced improvements by explicitly modeling similarity, using a Siamese network to track to improve localization and count accuracy.

Further enhancements in similarity modeling were proposed by Shi et al.~\cite{shi2022represent} and You et al.~\cite{you2023few}, who integrated advanced techniques such as self-attention and learnable similarity metrics to reduce appearance variability and guide feature fusion, respectively. Moreover, the incorporation of a vision transformer for feature extraction by Liu et al.~\cite{liu2022countr} marked a significant step towards leveraging the strengths of transformer models in the field, demonstrating the use of cross-attention to merge image and exemplar features effectively.
Most recently, LOCA~\cite{Dukic_2023_ICCV} introduces an object prototype extraction module that iteratively merges exemplar shape and appearance with image features, enabling adaptation from low-shot to zero-shot object counting scenarios.

However, these methods all need numerous training data, and the exemplar annotation is labor-consuming. In specific, the most commonly used dataset in CAC task, FSC-147~\cite{ranjan2021learning}, includes 3659 training images with more than 10,000 exemplar annotations needed.

Ma et al.~\cite{ma2023can} first address CAC with Segment Anything Model (SAM)~\cite{kirillov2023segany} in a training-free manner. 
% \hm{
After that Shi et al.~\cite{shi2023training} proposed a novel training-free object counter that introduces prior-guided mask generation to improve accuracy, along with a two-stage text-based object counting method, yielding competitive results on standard benchmarks. 
However, the performance of these training-free methods heavily lags behind the training-based CAC methods. In this work, we propose a simple but effective training-free approach to significantly reduce the performance gap.
% However, SAM underperforms compared to few-shot counting methods, particularly with small and congested objects. This is attributed to its tendency to segment clustered objects of the same category into one mask and the lack of semantic class annotations in its training masks, affecting its object differentiation ability.
% Nevertheless, as shown in Figure~\ref{fig:sam_vs_dino}, the above two issues still exist, especially the lack of semantic information.
% }
% \begin{figure}
% \centering
% \includegraphics[width=\linewidth]{images/bad.pdf}  
% \caption{Examples from Training-Free Object Counting~\cite{shi2023training}.}
% \label{fig:TRbad}
% \end{figure}

\begin{figure*}
\centering
\includegraphics[width=0.9\linewidth]{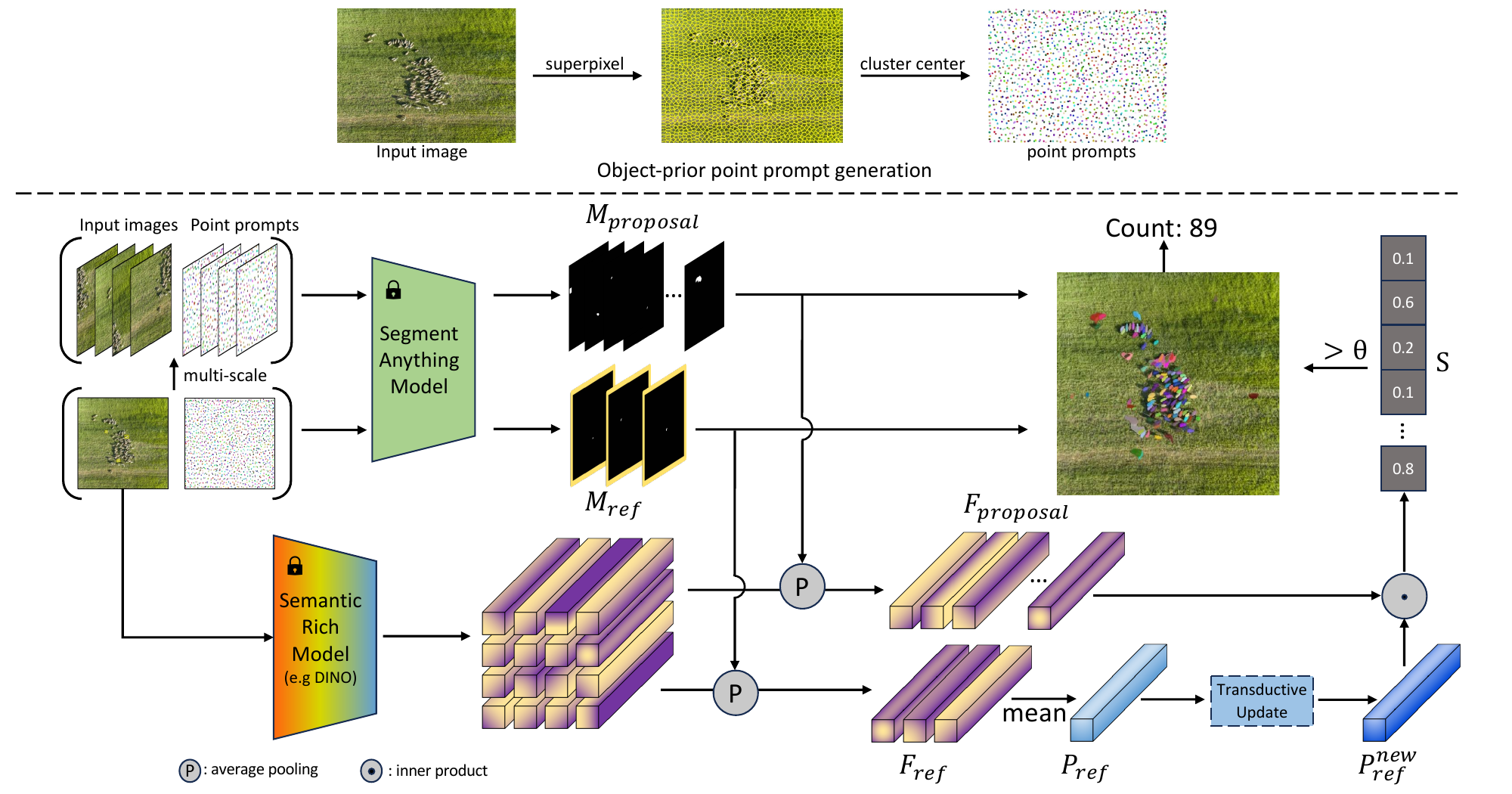} 
\caption{
% \textbf{Top} row shows the generation procedure of object-prior point prompts. \textbf{Bottom} row shows the pipeline overview of our approach. The transductive update module is presented in Eq.~\ref{eq:tpu}.
\textbf{Top} row illustrates the creation process of object-prior point prompts. \textbf{Bottom} row depicts a pipeline overview of our proposed training-free object counting approach. Details of the transductive update module are provided in Eq.~\ref{eq:tpu}. 
% Reference objects are marked in yellow boxes in input image (zoom in for a better view).
% In the input image, reference objects are highlighted in yellow boxes (zoom in for better visibility).
Reference objects are marked with yellow boxes in the input image (zoom in for clarity).
}
\label{fig:main}
\vspace{-0.5cm}
\end{figure*}
%-------------------------------------------------------------------------
\subsection{Superpixel Methods}
% say some superpixel methods, where are they usually used, why we need to use in this work
Superpixel is an important concept in computer vision, primarily used to group pixels into perceptually meaningful regions, which can significantly reduce the complexity of image processing tasks. As a way to effectively reduce the number of image primitives for subsequent processing, superpixel algorithms have been widely adopted in vision problems such as saliency detection~\cite{yang2013saliency}, object detection~\cite{gould2008multi}, and semantic segmentation~\cite{shu2013improving}.  

Superpixel was popularized by Ren and Malik~\cite{ren2003learning} through their pioneering work on the Normalized Cuts algorithm, and has been significantly advanced by the introduction of the Simple Linear Iterative Clustering (SLIC)~\cite{achanta2012slic} algorithm. SLIC uses K-means clustering in a five-dimensional space of color and pixel coordinates to create compact and uniform superpixels, which has made it a benchmark in the field.  Several variants(LSC~\cite{li2015superpixel}, Manifold SLIC~\cite{liu2016manifold}, SNIC~\cite{achanta2017superpixels}) have been developed upon SLIC.

% Building upon SLIC, several variants have been developed to cater to different segmentation needs. The Linear Spectral Clustering (LSC) algorithm~\cite{li2015superpixel} extends the feature space to ten dimensions and employs weighted K-means to improve boundary adherence and segmentation detail. Manifold SLIC~\cite{liu2016manifold} projects images onto a two-dimensional manifold, taking into account the image content to generate more sensitive superpixels. Meanwhile, Simple Non-Iterative Clustering (SNIC)~\cite{achanta2017superpixels} forgoes iterative K-means in favor of a non-iterative region-growing approach, which can enhance efficiency.

%-------------------------------------------------------------------------
\subsection{Vision Foundation Models}
% say some foundation models (SAM, CLIP, DINO,...), and what the difference of these methods, and why we need to use DINO

% The advancement of large-scale vison-language models such as CLIP~\cite{caron2021emerging} has not only revolutionized natural language processing but has also made a significant impact in the field of computer vision. Known as "foundation models," these robust systems excel at generalizing from limited examples, showcasing their strengths in zero-shot and few-shot learning scenarios. CLIP utilizes contrastive learning to jointly train text and image encoders, which allows CLIP to understand and categorize new visual concepts using text prompts, demonstrating remarkable zero-shot learning abilities across a variety of visual domains.

For pure vision area, DINO (self-DIstillation with NO labels)~\cite{caron2021emerging} and its more advanced iteration, DINOv2~\cite{oquab2023dinov2}, represent significant strides in self-supervised learning within computer vision. Through a teacher-student distillation process that does not rely on labeled data, DINO masters semantic-rich visual representation. DINOv2 builds on this by refining the architecture and training mechanisms, further improving the model's efficiency and the accuracy of learned representations.

Most recently, SAM~\cite{kirillov2023segany} specializes in image segmentation, adeptly generating accurate object masks from prompts such as points and boxes. SAM's proficiency is evidenced by its remarkable performance on diverse segmentation benchmarks and its zero-shot transfer abilities, which extend across a range of datasets~\cite{chen2023anydoor,deng2023segment}.

These foundation models collectively redefine the capabilities of computer vision systems. They provide a versatile framework for tackling new tasks and adapting to unfamiliar data distributions without the necessity of bespoke training~\cite{yang2023track,ma2023segment,wang2023caption}.

%-------------------------------------------------------------------------
\section{Method}
\label{sec:Method}

% \hm{add overall pipeline description here, you can refer to Sec3 of https://arxiv.org/pdf/2203.08354.pdf}
% In this section, we present four novel modules to improve the performance of using SAM for training-free CAC, i.e., object-prior point prompt (Sec.~\ref{subsec:object-prior}) and semantic-rich feature (Sec.~\ref{subsec:semantic-feat}) are matters for the training-free model to achieve a decent counting result. Multi-scale strategy (Sec.~\ref{subsec:multi-scale}) and transductive prototype construction (Sec.~\ref{subsec:transductive-prototype}) are helpful to significantly boost the counting accuracy. An overall framework of our approach is shown in Fig.~\ref{fig:main}.
In this section, we introduce four innovative modules that enhance the effectiveness of employing SAM in training-free CAC, i.e., object-prior point prompt (Sec.~\ref{subsec:object-prior}) and semantic-rich feature (Sec.~\ref{subsec:semantic-feat}) are matters for the training-free model to deliver satisfactory counting outcomes. Additionally, the multi-scale strategy (Sec.~\ref{subsec:multi-scale}) and transductive prototype updating (Sec.~\ref{subsec:transductive-prototype}) help substantially elevate counting accuracy. The comprehensive framework of our method is depicted in Figure~\ref{fig:main}.

\subsection{Object-prior Point Prompt Matters}\label{subsec:object-prior}
\begin{figure}
\centering
\includegraphics[width=\linewidth]{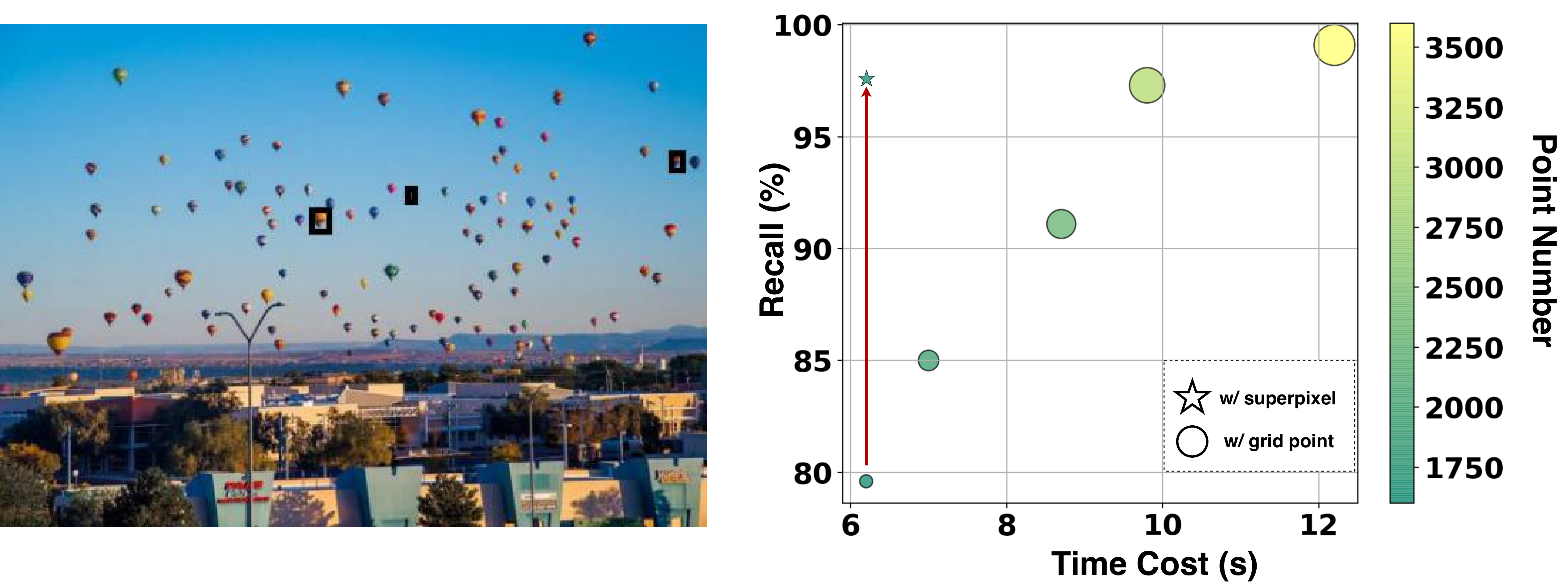}  
\caption{Effectiveness demonstration of the use of superpixel on the quality of mask proposals generated by SAM. \textbf{Left:} an image with reference exemplars (hot-air balloon) in black boxes. \textbf{Right:} a bubble chart illustrates the trade-off between the recall rate of the interested object and the time cost. Without increasing the number of point prompts, SAM with superpixel\protect\footnotemark can significantly improve the recall rate of hot-air balloons in the output mask proposals, thus avoiding the potential computational burden caused by the demand for denser point grids. Please find visualizations of the generated mask proposal in the appendix.}
% It clearly shows that 
% SAM with superpixel can achieve a high recall rate while maintaining a lower time cost compared to the default point grid prompts.  
% utilizing superpixel for SAM significantly enhances the interested object recall rate without adding to the quantity of point prompts, effectively bypassing any increase in computational load.
% \hm{total number is 113, recall 1600: (113-23)/113 = 79.6\% (miss 23), 3600: 112/113=99.1\% (miss 1), superpixel: 110/113=97.3\% (miss 3)? time cost? sp:10ms, 40x40:6212ms, 60x60:12294ms}
\label{fig:superpixel_vs_normal}
\vspace{-0.4cm}
\end{figure}

In the use of SAM for zero-shot object proposal generation~\cite{kirillov2023segany}, a regular grid of point prompts is utilized to automatically generate class-agnostic masks. In the scenario of object counting, counted objects are usually tiny or crowded. In order to increase the recall of the object proposals, fine-grained grid point prompts became indispensable. As presented in Figure~\ref{fig:superpixel_vs_normal}, \hm{naively increasing the number of points grid from 1600 to 3600 increases recall by 20\%}. However, 
an increase in the number of point grids leads to a doubling of computational time overhead. 
In order to ensure the recall rate without increasing time complexity too much, we propose to utilize the superpixel segmentation algorithm~\cite{ren2003learning} to firstly perceptual group pixels into clusters and then take cluster centers as object-prior point prompts to feed into SAM for object proposal generation.

In this work, a superpixel algorithm named Simple Linear Iterative Clustering (SLIC) is used to generates compact superpixels with a low computational overhead~\cite{achanta2012slic}. 
% In SILC, an initial number of cluster centers is required. Following the existing training-free CAC method~\cite{}, we use an adaptive number $K$ for the object-counting task
% \begin{equation}
%     K = \lambda \times \frac{S_I}{\hat{S_r}}
% \end{equation}
% where $S_I$ is the area of the image, $\hat{S_r}$ is the average area of the given few-shot reference boxes, and \hm{$\lambda$ is the density hyperparameter which is set to ? in our study}.
\footnotetext{Performing superpixel on this example image takes only 10ms with AMD Ryzen 9 3900X CPU, a negligible computational expense compared to the SAM computation cost.}

Given the initial cluster number $K$, SLIC can directly output superpixel centers $\mathcal{C} = \{(x_i, y_i)\}_{i=1}^{K}$.
% which can then be fed into SAM together with the provided reference boxes (or points) $\mathcal{R}$ batch by batch for mask proposal generation
These centers, along with the provided reference boxes (or points) $\mathcal{R}$, can then be fed into SAM batch by batch for mask proposal generation
\begin{equation}
    \mathcal{M}_\text{ref} = \mathbb{D}(\mathbb{E}(I), \mathcal{R}), \quad \mathcal{M}_\text{proposal} = \mathbb{D}(\mathbb{E}(I), \mathcal{C})
\end{equation}
where $\mathbb{E}$ and $\mathbb{D}$ are encoder and decoder of SAM respectively. $I$ is the input image. $\mathcal{M}_\text{ref}$ denotes the generated masks for corresponding referenced prompts \textcolor{black}{(reference boxes or points)} and $\mathcal{M}_\text{proposal}$ means all potential mask proposals~\footnote{\textcolor{black}{
% The mask for background 
% The outlier masks 
The largest mask generated by SAM, which typically corresponds to the background, 
and masks for referenced objects are filtered from $\mathcal{M}_\text{proposal}$ for simplicity, i.e., $\mathcal{M}_\text{proposal} \cap \mathcal{M}_\text{ref} = \emptyset$.}}.

\subsection{Semantic-rich Feature Matters}\label{subsec:semantic-feat}
When object masks are ready, the next step is to recognize all object masks similar to the given reference objects from $\mathcal{M}_{\text{proposal}}$. One straightforward way is to compare the similarity of the region-of-mask features (RoMF) of referenced objects to that of the RoMF of others
\begin{align}
    & \mathcal{S}^i = \text{sim}(\mathsf{P}_{\text{ref}}, \ F^i_{\text{proposal}}), \quad \text{where} \\ 
    & \mathsf{P}_{\text{ref}} = \frac{1}{n_\text{ref}} \sum_{j=1}^{n_\text{ref}} \text{AvgPool}\left (\mathbb{E}(I),\ \mathcal{M}^j_{\text{ref}}\right ) \nonumber \\ 
    & F^i_{\text{proposal}} = \text{AvgPool}\left (\mathbb{E}(I), \ \mathcal{M}^i_{\text{proposal}}\right ) \nonumber
\end{align}
where $\text{AvgPool}$ is the average pooling operation. $n_{\text{ref}}$ is the number of given reference objects and is set to 3 by default in the literature~\cite{ranjan2021learning}. $\mathsf{P}_\text{ref}$ denotes the average of referenced object features, a.k.a prototypes. $\text{sim}(\cdot, \cdot)$ is the similarity function and the cosine function is used in this study. $\mathcal{S}^i$ is the similarity score for each object proposal $i$.

\begin{figure}
\centering
\includegraphics[width=\linewidth]{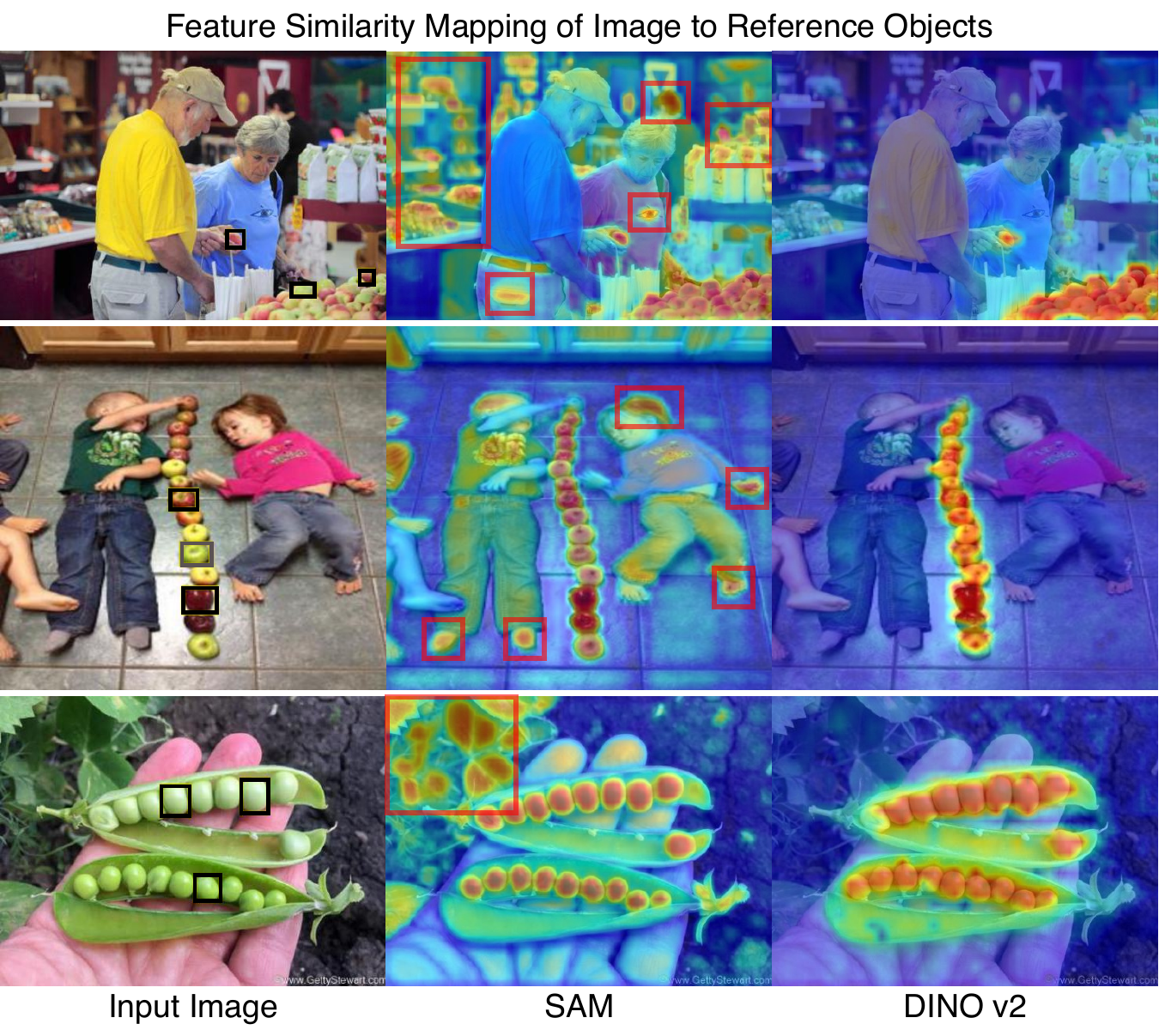}  
\caption{
% Examples from Training-Free Object Counting~\cite{shi2023training}.And the correspond similarity map from SAM and DINO
% Visualization of similarity mapping of image to reference object (in black box) features of SAM and DINO v2. It clearly shows that the similarity mapping of SAM miss highlights many areas not belonging to the interested object but the similarity mapping of DINO v2 precisely cover interested objects. 
Visualization of the similarity mappings of image features to reference object features (marked in black boxes) using SAM and DINOv2. It distinctly shows that SAM's similarity mapping erroneously highlights numerous areas unrelated to the target object, whereas DINOv2's mapping accurately encompasses the objects of interest. This demonstrates that DINOv2's features possess more semantically relevant knowledge.
}
\label{fig:sam_vs_dino}
\vspace{-0.5cm}
\end{figure}

Next, an overall segmentation for interested objects can be obtained by selecting mask proposals whose similarity score is higher than a predefined threshold $\theta$. \hm{However, searching for an optimal $\theta$ can be hard} due to the feature representations of SAM lacking clear semantic-related knowledge~\cite{ma2023can,li2023semantic}. As empirical analysis presented in Figure~\ref{fig:sam_vs_dino}, many objects of different categories from the reference object may have a quite close similarity score to those objects of the same category, thus it is inevitable to output poor counting results with noisy segment generation. \hm{For instance, the SAM feature struggles to differentiate between apples and other messy objects in row 1, as well as children's hair and feet in row 2 of Figure~\ref{fig:sam_vs_dino}.
% mess objects in background vs apples in row 1, children's hair and feet vs apple in row 2, leaves vs beans in row 3
}
% Moreover, we empirically find that the increased recall for referenced objects achieved through the proposed superpixel method fails to benefit the final counting performance when directly using SAM feature representation for object discrimination.
Moreover, our empirical findings (the first two rows in Table~\ref{tab:aba}) indicate that the enhanced recall of referenced objects achieved through the proposed superpixel method does not improve the final counting performance when the SAM feature representation is directly employed for object discrimination. Therefore, to obtain a plausible counting accuracy, using semantic-rich features for mask proposal selection becomes indispensable. 
\begin{figure*}[!ht]
\centering
\includegraphics[width=\linewidth]{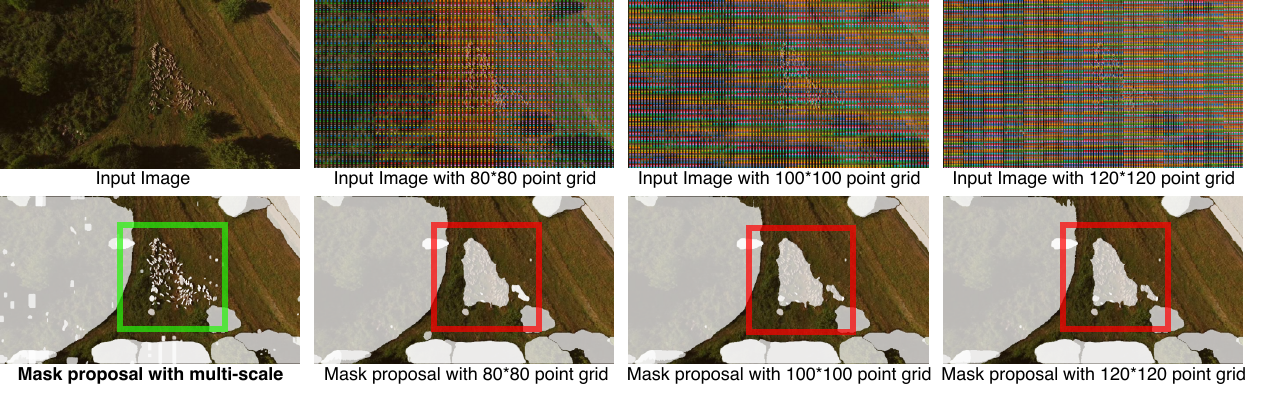}  
% \vspace{-0.3cm}
\caption{ 
% \hm{$32\times32$ points are used for images in our multi-scale mechanism.}
% Effectiveness demonstration of our proposed multi-scale mechanism for scenarios with extremely tiny counting objects. Simply increasing the number of point prompts does not allow SAM to generate precise instance-level mask proposals but SAM with our multi-scale mechanism (32*32 points are used in this demonstration) successfully fulfills the target. 
Effectiveness demonstration of our multi-scale mechanism in scenarios involving extremely tiny counting objects. Merely increasing the quantity of point prompts\protect\footnotemark in SAM fails to yield accurate instance-level mask proposals. However, integrating SAM with our multi-scale mechanism (using 32*32 points for aerial photography of sheep in this demonstration) effectively achieves better mask quality.
}
\label{fig:dense-grid-fail}
\vspace{-0.5cm}
\end{figure*}
\footnotetext{Multiple colors are used for the point grid in the figures of columns 2-4 for better visualization.}

Recently, many vision foundation models with rich semantic knowledge have been made accessible to the public, such as DINO~\cite{caron2021emerging,oquab2023dinov2}.

% and CLIP~\cite{caron2021emerging}. 
Combining with the semantic-rich feature representation, the final count can be obtained
\begin{align}
    &\text{count} = n_{\text{ref}} + \sum_{i=1}^{N} \mathbbm{1}(\mathcal{S}^i > \theta), \quad \text{where} \\
    & \mathcal{S}^i = \text{sim}(\mathsf{P}_{\text{ref}}, \ F^i_{\text{proposal}}), \nonumber \\ 
    & \mathsf{P}_{\text{ref}} = \frac{1}{n_\text{ref}} \sum_{j=1}^{n_\text{ref}} \text{AvgPool}\left (\mathbb{E}_{\text{sem}}(I),\ h(\mathcal{M}^j_{\text{ref}}) \right ) \nonumber \\ 
    & F^i_{\text{proposal}} = \text{AvgPool}\left (\mathbb{E}_{\text{sem}}(I), \ h(\mathcal{M}^i_{\text{proposal}})\right ) \nonumber
\end{align}
where $\mathbb{E}_{\text{sem}}$ is the encoder with rich semantic knowledge and $h(\cdot)$ is the interpolation function to accommodate the resolution of SAM generated mask to that of the encoder output. $N$ is the number of object proposals. $\theta$ is the similarity threshold and \hm{is empirically set to 0.4 in this study}.

\subsection{Multi-scale Segmentation Helps}\label{subsec:multi-scale}
In the task of object counting, there exist some scenes where objects that need to be counted are extremely tiny and crowded, e.g., aerial photography of sheep. It becomes challenging for SAM to generate reasonable instance-level masks. As Figure~\ref{fig:dense-grid-fail} shows, naive increasing the number of points from 6400 to even 14400 still fails to make SAM generate precise mask proposals for the sheep instance. In order to tackle this problem, we propose a simple multi-scale mechanism. Specifically, we first cut the input image into $n_{\text{p}}\times n_{\text{p}}$ patches of the same size\footnote{Empirically, $n_{\text{p}}=2$ is found to perform well in this study.}. Then we resize these patches to the original size of the input image and feed them together with the original input image into SAM for mask proposal generation. Please see Figure~\ref{fig:main} for details. As the bottom left picture in Figure~\ref{fig:dense-grid-fail} shows, using this simple multi-scale mechanism can successfully obtain precise instance-level mask proposals.

% cost comparison

\begin{table*}[t]
\centering
\resizebox{0.8\textwidth}{!}{%
\begin{tabular}{lcccccc}
\toprule
\multicolumn{1}{c}{} & \multirow{2}{*}{\textbf{Training}} &\multirow{2}{*}{\makecell[c]{\textbf{Reference
} \\ \textbf{Format}}} & \multicolumn{2}{c}{\textbf{FSC-147}} & \multicolumn{2}{c}{\textbf{CARPK}} \\
\cmidrule(lr){4-5} \cmidrule(lr){6-7}
 &        &        & \textbf{MAE $\downarrow$}   & \textbf{RMSE $\downarrow$}    & \textbf{MAE $\downarrow$} & \textbf{RMSE $\downarrow$}  \\ 
 \hline
GMN~\cite{lu2019class} {\tiny [ACCV 19'] }  & Yes  & Box     &26.52    &124.57       &9.90    &-       \\
FamNet~\cite{ranjan2021learning}  {\tiny [CVPR 21'] }   & Yes  & Box  &22.08    &99.54   &18.19    &33.66        \\
CFOCNet+~\cite{yang2021class}  {\tiny [WACV 21'] }     & Yes & Box &22.10    &112.71   &-    &-       \\
BMNet+~\cite{shi2022represent}  {\tiny [CVPR 22'] }   & Yes & Box &14.62    &91.83    &5.76     &7.83        \\
SAFECount~\cite{you2023few} {\tiny [WACV 23'] }     & Yes & Box &14.32    &85.54   &5.33     &7.04       \\
\textcolor{black}{PseCo}~\cite{huang2024point} {\tiny [CVPR 24'] }   & \textcolor{black}{Yes}     & \textcolor{black}{Box}    &\textcolor{black}{13.05}    &\textcolor{black}{112.86}         &-    &-      \\
LOCA~\cite{Dukic_2023_ICCV}   {\tiny [ICCV 23'] }     & Yes  & Box      &10.79    &56.97     &9.97   &12.51        \\\hline
SAM Baseline~\cite{shi2023training}    & No     & N.A.    &42.48    &137.50          &16.97    &20.57        \\
Count-Anything~\cite{ma2023can}    & No      & Box   &27.97    &131.24          &-    &-        \\
TFOC~\cite{shi2023training} {\tiny [WACV 24'] }   & No     & Box    &19.95    &132.16         &10.97    &14.24       \\
% \textcolor{blue}{PseCo}~\cite{huang2024point} {\tiny [CVPR 24'] }   & \textcolor{blue}{No}     & \textcolor{blue}{Box}    &\textcolor{blue}{13.05}    &\textcolor{blue}{112.86}         &-    &-      \\

\rowcolor{Gray}

\textbf{Ours}    & No   & Box      &\textbf{12.26}   &\textbf{56.33}       &\textbf{4.39}    &\textbf{5.70}       \\\hline
TFOC~\cite{shi2023training}{\tiny [WACV 24'] }   & No  & Point  &20.10 & 132.83   &11.01   & 14.34         \\
\rowcolor{Gray}
\textbf{Ours} & No  & Point &\textbf{12.47}   &\textbf{49.97}      &\textbf{4.39}    &\textbf{5.70} \\\hline
TFOC~\cite{shi2023training}{\tiny [WACV 24'] }   & No  & Text  &24.79 & 137.15  &-   & -        \\
\rowcolor{Gray}
\textbf{Ours} & No  & Text &\textbf{23.59}  &\textbf{113.60}      &-  &- \\

% TFOC~\cite{shi2023training} {\tiny [WACV 24'] } &No&24.79 &137.15\\
% \rowcolor{Gray}
% \textbf{Ours} &No&\textbf{23.59} &\textbf{113.60}\\\hline
\bottomrule
\end{tabular}
}
\caption{
% Effect of our approaches with point and box reference format on FSC-147 and CARPK benchmarks. Our method outperforms the training-free methods and demonstrates competitive performance compared to training-based approaches. The bold font highlights the best counting results among training-free methods.
Quantitative comparisons between our methods and others using point and box reference formats on the FSC-147 and CARPK benchmarks are presented. Our approach surpasses other training-free methods and shows competitive results against training-based ones. The best counting outcomes among training-free methods are emphasized in bold font.
% , while the underlined font indicates the best counting results among learning-based methods. 
% This convention is consistent throughout the tables presented below. 
% \hm{1. add one column named: reference format; 2:another two evaluation metric can be listed if ours can be better}
}
\label{tab:main} 
\vspace{-0.5cm}
\end{table*}

\subsection{Transductive Prototype Updating Helps}\label{subsec:transductive-prototype}
In Sec.~\ref{subsec:semantic-feat}, the prototype of the interested object $\mathsf{P}_{\text{ref}}$ is simply the average of given few-shot reference object features. Since objects in the same class may have various appearances in the same image, this simple prototype may not be representative enough to match all interested objects. We argue that potentially interested-object features from mask proposals can be helpful for matching generalization and propose a novel transductive prototype updating strategy to improve the quality of the prototype
\begin{equation}\label{eq:tpu}
    \mathsf{P}_{\text{ref}}^{\text{new}} = \frac{n_\text{ref} \cdot \mathsf{P}_{\text{ref}} + \sum_i^N \mathbbm{1}(\mathcal{S}^i > \delta)\cdot F^i_{\text{proposal}}}{n_\text{ref} + \sum_i^N \mathbbm{1}(\mathcal{S}^i > \delta)}
\end{equation}
where $\delta$ is a similarity threshold and is set to 0.5 in this study for selecting plausible reference candidates.

\section{Experiment}
\label{sec:Experiment}
\begin{figure*}
\centering
\includegraphics[width=0.9\linewidth]{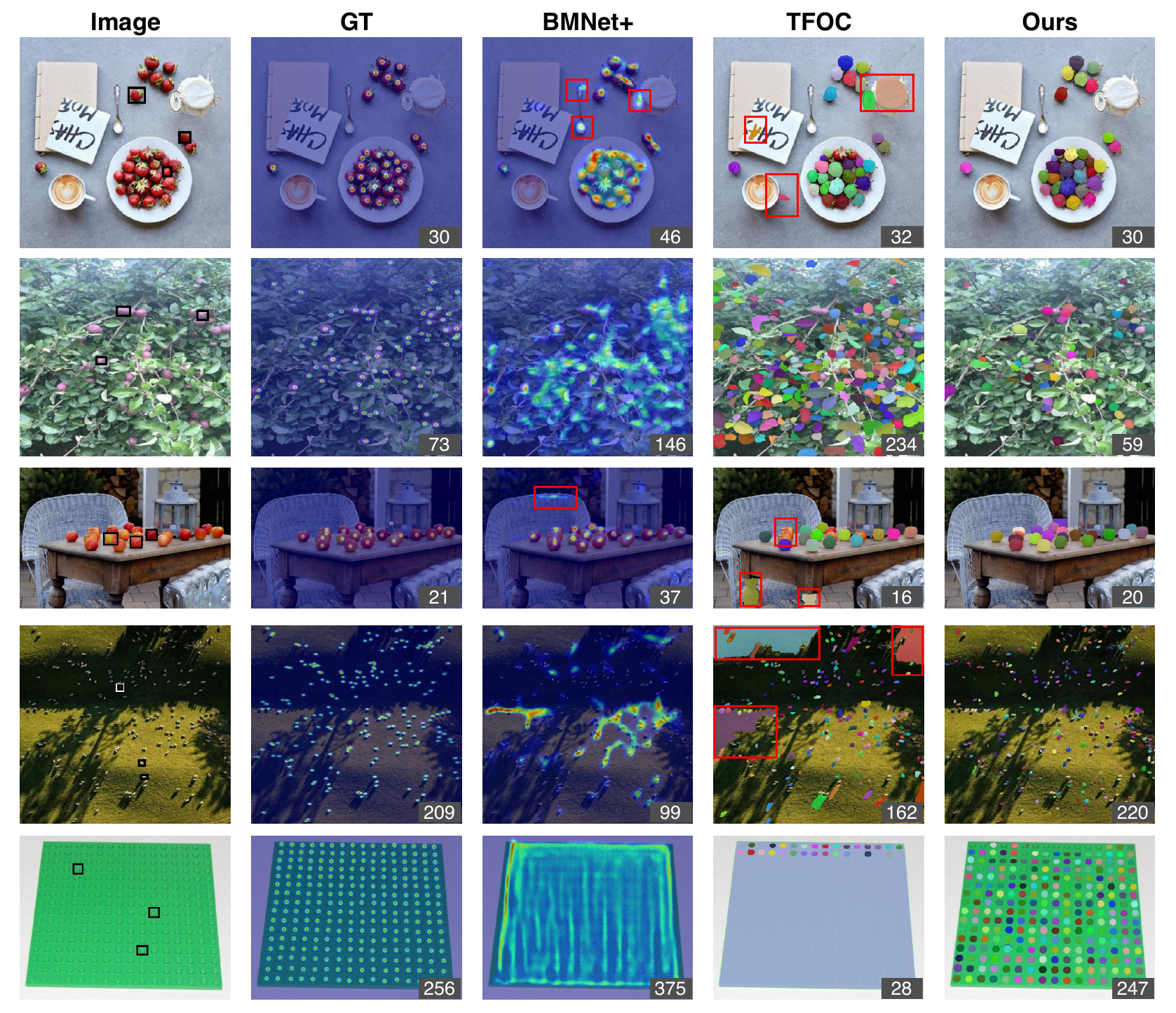}  
\caption{
% Qualitative results on the FSC-147 dataset. 
% The samples on the left exhibit significant intra-class variations such as scale, pose, and illumination condition. 
% The black (white) boxes indicate exemplars. Counting values are shown at the bottom-right corner. Our method can predict accurate masks in both dense and sparse scenes. Best viewed by zooming in.
Qualitative results on the FSC-147 dataset are displayed. Exemplars are marked with black (white) boxes, and counting values are noted in the bottom-right corner. Error predictions are indicated by red boxes. Our method accurately predicts masks in both densely and sparsely populated scenes. Best viewed by zooming in.
}
\label{fig:quality}
\vspace{-0.5cm}
\end{figure*}

In this section, we first provide details of our experimental setup. Next, we assess our proposed method on two widely-used CAC benchmarks, comparing it against both training-based and training-free CAC methods. Lastly, we conduct comprehensive ablation studies to further explore the effectiveness of our approach.
%-------------------------------------------------------------------------
\subsection{Experimental Setup}
\textbf{Dataset:} 
% In our experiment, we evaluate our approach on the FSC-147 and CARPK, which are two commonly used datasets in the counting task. The FSC-147 is the first large-scale CAC dataset, which contains 6,135 images from 147 distinct categories. The FSC-147 is splited into training, validation and test set, which consists of 3659,1286 and 1190 images, respectively. The annotation of each image includes three exemplars with their bounding boxes and the central points. As our approach does not have training progress, we only use the test set, which  include 1,190 images from 29 categories. In the CARPK dataset, it contains 1,448 car park images from drone view. We follow the same settings in~\cite{Dukic_2023_ICCV,shi2022represent,shi2023training},twelve exemplars are sampled from the training CARPK images and used in all test CARPK images. Same as FSC-147, we only evaluate on the testset, which consists of 459 images.
In our study, we assess our method using two popular datasets for counting tasks: FSC-147~\cite{ranjan2021learning} and CARPK~\cite{hsieh2017drone}. FSC-147, the first extensive CAC dataset, includes 6,135 images across 147 categories, divided into 3,659 training, 1,286 validation, and 1,190 test images. Each image is annotated with bounding boxes and central points of three exemplars. Since our approach doesn't require training, we focus solely on the test set, which features 1,190 images from 29 categories. The CARPK dataset comprises 1,448 images of car parks captured from drones. Following the protocol in~\cite{Dukic_2023_ICCV,shi2022represent,shi2023training}, we select twelve exemplars from the training images of CARPK for use in all test images. Similar to FSC-147, our evaluation is confined to the test set, containing 459 images.

% \smallskip
\noindent \textbf{Evaluation Metrics:} Follow \cite{Dukic_2023_ICCV,shi2022represent}, we report the Mean Absolute Error (MAE), Root Mean Square Error (RMSE). In particular, MAE = $\frac{1}{n} \sum^{n}_{i=1}|y_i-\hat{y_i}|$, RMSE = $\sqrt{\frac{1}{n} \sum^{n}_{i=1}(y_i-\hat{y_i})^2}$,  where n is the number of test images, $y$ is the ground truth and $\hat{y}$ is the prediction.

%-------------------------------------------------------------------------
% \smallskip
\noindent \textbf{Comparing Methods:} 
% We compare our method with current state of the art approaches in both training and training-free scenarios. In training scenarios, we compare our results with following CAC methods: GMN (General Matching Network~\cite{lu2019class}), FamNet (Few-shot adaptation and matching Network~\cite{ranjan2021learning}), CFONet+ (Class-agnostic Fewshot Object Counting Network~\cite{yang2021class}), and most recent state of the art approaches BMNet+(Bilinear Matching Network~\cite{shi2022represent}), SAFECount(similarity-aware feature enhancement~\cite{you2023few}) and LOCA(Low-shot Object Counting network with iterative prototype Adaptation~\cite{Dukic_2023_ICCV}). In training-free setting, we compare our models with two available training-free CAC methods: Count-Anything~\cite{ma2023can} and TFOC (training-free object counting~\cite{shi2023training}).
Our approach is benchmarked against the latest state-of-the-art methods in both trained and training-free scenarios. For trained scenarios, we compare with several CAC techniques: GMN (General Matching Network~\cite{lu2019class}), FamNet (Few-shot adaptation and matching Network~\cite{ranjan2021learning}), CFONet+ (Class-agnostic Fewshot Object Counting Network~\cite{yang2021class}), the advanced BMNet+ (Bilinear Matching Network~\cite{shi2022represent}), SAFECount (similarity-aware feature enhancement~\cite{you2023few}), and LOCA (Low-shot Object Counting network with iterative prototype Adaptation~\cite{Dukic_2023_ICCV}). In the training-free setting, our models are evaluated alongside two existing methods: Count-Anything~\cite{ma2023can} and TFOC (training-free object counting~\cite{shi2023training}).

All experiments in this work are conducted on one NVIDIA GeForce RTX 3090 GPU and all other experimental settings (\textcolor{black}{including the text prompt}) are identical to the TFOC~\cite{shi2023training} approach for a fair comparison.
% \textcolor{blue}{ On both dataset (FSC-147 and CARPK), the average inference time is 5.32s per image for 2500 (50*50) initial points.}
% The number of clusters in SLIC is simply $\frac{Area_{img}}{Area_{patch}}$. 

% . We utilize the “vit\_h” image encoder in SAM and "vits14" in DINO v2. To incorporate
% box prompts, we use annotated reference object boxes
% provided by the dataset and extract their center points as
% point prompts. \hm{We dynamically set as t=($S_{img}\| S_{ref}$) * 9, where $S_{ref}$
% represents the minimum size of reference objects obtained
% from their masks, and $\|$ denotes exact division. TO BE REVISED}
% \subsection{Computational resource:}

\subsection{Comparison with State-of-the-Arts}
% \subsubsection{Quantitative Results:}
\textbf{Quantitative Results:}
Table~\ref{tab:main} reports the comparison results on two popular CAC benchmarks. First, our approach achieves consistent performance improvements over the compared training-free methods on various evaluation metrics. Specifically, On the FSC-147 dataset, our approach lowers the Mean Absolute Error (MAE) to 12.26, outperforming the naive SAM Baseline, which has an MAE of 42.48, and the Count-Anything method with an MAE of 27.97. Compared to the state-of-the-art training-free method TFOC~\cite{shi2023training}, the performance gain of our approach reaches \textbf{+7.69} MAE on the FSC-147 dataset and \textbf{+6.58} MAE on CARPK dataset. Meanwhile, the RMSE metric of our approach is also significantly lower than the TFOC method which demonstrates that our method can perform better in scenarios with more crowded objects.

Second, our approach greatly reduces the performance gap between the training-based and training-free methods. For example, on FSC-147, the gap has decreased from 9.16 to 1.47 on MAE and even outperformed 0.64 on RMSE. On the CARPK dataset, our approach excels in all the compared algorithms and creates a state-of-the-art performance.

\begin{table}[!t]
\centering
\resizebox{0.95\columnwidth}{!}{
\begin{tabular}{@{}lccccccccccccc@{}}
\toprule
&\multirow{2}{*}{\textbf{Training}} & \multicolumn{4}{c}{\textbf{FSC-147}}\\
\cmidrule(lr){3-6}
& & \textbf{MAE} $\downarrow$ &
\textbf{RMSE} $\downarrow$\\
\hline
FamNet~\cite{ranjan2021learning}  {\tiny [CVPR 21'] }  &Yes &26.76 &110.95 \\
BMNet+~\cite{shi2022represent}  {\tiny [CVPR 22'] }  &Yes &16.89 &96.65 \\
LOCA~\cite{Dukic_2023_ICCV}   {\tiny [ICCV 23'] } &Yes &\textbf{12.53} &75.32\\\hline
TFOC~\cite{shi2023training} {\tiny [WACV 24'] }  &No &22.96 &137.27 \\
\rowcolor{Gray}
\textbf{Ours} &No&15.15 &\textbf{69.16}\\\hline

% \bottomrule
\end{tabular}
}
\caption{Evaluation on one-shot setting on FSC-147 test set.}
\label{tab:one_shot} 
\vspace{-0.5cm}
\end{table}
% \begin{table}[h]
% \centering
% \resizebox{0.35\textwidth}{!}{
% \begin{tabular}{l|cc}
% \toprule
% Method & MAE $\downarrow$   & RMSE $\downarrow$  \\
% \hline
% % GMN\cite{lu2019class}  &26.52 &124.57 \\
% FamNet~\cite{ranjan2021learning}  {\tiny [CVPR 21'] }  &26.76 &110.95 \\
% BMNet+~\cite{shi2022represent}  {\tiny [CVPR 22'] }  &16.89 &96.65 \\
% LOCA~\cite{Dukic_2023_ICCV}   {\tiny [ICCV 23'] }  &12.53 &75.32\\
% TFOC~\cite{shi2023training} {\tiny [WACV 24'] }   &22.96 &137.27 \\
% \hline

% \textbf{Ours}  &15.15 &69.16\\
% % \hline
% \bottomrule
% \end{tabular}
% }
% \caption{Evaluation on one-shot setting on FSC-147 testset.}
% \label{tab:one_shot} 
% \end{table}

Finally, we also evaluate the proposed method with point references and the consistent superior performance demonstrates the robustness of our approach.

We also evaluate our pipeline on FSC-147 test set with a one-shot counting scenario. Table~\ref{tab:one_shot} indicates that our approach is robust on a one-shot counting scenario, which is consistent with the results of the three-shot above.

And for text prompts, we followed TFOC~\cite{shi2023training} and employ the CLIP-Surgery model to derive coarse bounding boxes for the target object using text references. Then our approach use these coarse bounding boxes for object counting. As presented in last rows of the Table \ref{tab:main}, our approach is sightly superior to the compared methods. The main reason is that the bounding boxes generated from TFOC~\cite{shi2023training} are not accurate. Technically, great improvement will be observed if high quality bounding boxes are given.
% compare to training-free methods, especially tfoc
% compare to training-based methods
% point

% As shown in Table~\ref{tab:main}, our approach achieve significant improvement over the most recent training-free SOTA, which is 38.54\% and 55.96\% on FSC-147 and CACRPK dataset respectively. 

% Compared with training-based methods, our approach exhibits advantage over all of the training-based SOTA methods  on CARPK. With a improvement of and 17.6\% over the best performed SAFECount, sets a solid new state-of-the-art.

% On FSC-147, our approach also beat most of methods (FamNet, GMN, CFOCNet, BMNet+, and SAFECount),  compared with SAFECount and BMNet+, our approach achieves a relative improvement of 12.9\%  and 14.7\% respectively on FSC-147 test MAE. 
% \begin{table}[!t]
% \centering
% \resizebox{0.95\columnwidth}{!}{
% \begin{tabular}{@{}lccccccccccccc@{}}
% \toprule
% &\multirow{2}{*}{\textbf{Training}} & \multicolumn{4}{c}{\textbf{FSC-147}}\\
% \cmidrule(lr){3-6}
% & & \textbf{MAE} $\downarrow$ &
% \textbf{RMSE} $\downarrow$\\
% \hline
% % Xu \etal \cite{xu2023zero} {\tiny [CVPR 23']}&Yes&\underline{22.09} &\underline{115.17} \\
% SAM &No&42.48 &137.50 \\
% TFOC~\cite{shi2023training} {\tiny [WACV 24'] } &No&24.79 &137.15\\
% \rowcolor{Gray}
% \textbf{Ours} &No&\textbf{23.59} &\textbf{113.60}\\\hline

% \bottomrule
% \end{tabular}
% }
% \caption{Effect of our approaches with text prompt on FSC-147.}
% \label{tab:text-prompt}
% \vspace{-0.5cm}
% \end{table}
% \smallskip
% \subsubsection{Qualitative Analysis:}
\noindent \textbf{Qualitative Analysis:}
Figure~\ref{fig:quality} displayed the visualization results of various counting methods in different scenarios and our approach can always more accurately identify objects of interest in various scenes. Specifically, for scenes in the first three rows, both of the compared training-based and training-free methods mistakenly identified areas that do not belong to the interested object, such as the teacup handle and the spoon in row 1, the crowded leaves in row 2 and the table legs and the top edge of the chair back in row 3. While our approach can successfully and accurately segment out the interested object and produce a more accurate counting result. This will be attributed to the utilization of semantic-rich feature representations for SAM's mask proposal selection in our approach. Furthermore, for more challenging scenes (the last two rows), the density maps generated by the training-based BMNet+ method are vague and exhibit substantial deviations in counting outcomes.
% while the TFOC method either mistakenly counts large background areas as interested objects, or fails to recognize a large amount of crowded objects. 
The TFOC method either erroneously includes extensive background areas in its object count or overlooks numerous objects in crowded scenes.
% However, our approach can produce precise segmentation masks for each object instance and the output counting numbers are closer to the ground-truth ones thanks to the multi-scale mechanism of our approach.
Yet, our method yields accurate segmentation masks for individual objects due to its multi-scale mechanism, resulting in output counting numbers that more closely align with the ground truth.

\section{Ablation study}\label{sec:ablation-study}
\begin{table}[]
\centering
\resizebox{0.85\columnwidth}{!}{
\begin{tabular}{cccc|cc}
\toprule
\multicolumn{4}{c}{Components}  & \multicolumn{2}{c}{FSC-147}    \\ 
\cline{1-4} \cline{5-6} 
SP  & DINOv2 & TPU & MS & MAE $\downarrow$ & RMSE $\downarrow$ \\ 
\hline
-               &-            &-             &-            & 30.91 & 140.10 \\
\checkmark      & -           &  -           &   -         & 30.89 & 140.06  \\
-               & \checkmark  &  -           &   -         & \hm{20.19} & \hm{134.22}  \\
\checkmark      & \checkmark  &   -           &   -        & 17.32 & 132.64  \\
\checkmark      & \checkmark  &\checkmark      &   -       & 16.04 & 132.40 \\          
\checkmark      & \checkmark  &-  &   \checkmark           & 13.77  & 87.40   \\
\checkmark      & \checkmark  &\checkmark   &  \checkmark  & 12.26  & 56.33  \\ 
% \hdashline
% \checkmark     &    -          &\checkmark   & \checkmark  & \hm{15.47}  & 99.73  \\ 
% \checkmark     &    -          &\checkmark   & \checkmark  & 31.03  & 101.89  \\ 
\bottomrule
\end{tabular}}
\caption{Performance with different components on FSC-147. SP: superpixel, TPU: transductive prototype updating, MS: multi-scale.}
\label{tab:aba} 
\vspace{-0.3cm}
\end{table}

% \begin{table}[]
% \centering
% \begin{tabular}{cccc|cc}
% \toprule
% \multicolumn{3}{c}{Components}  & \multicolumn{2}{c}{CARPK}    \\ 
% \cline{1-4} \cline{5-6} 
% SP  & DINOv2 & TPU & MAE $\downarrow$ & RMSE $\downarrow$ \\ 
% \hline
% -               &-            &-                         & 14.27 & 16.61 \\
% \checkmark      & -           &  -                    & 19.63 & 21.90  \\
% -               & \checkmark  &  -                    & \hm{6.06} & \hm{8.51}  \\
% \checkmark      & \checkmark  &   -                   & 5.41 & 7.71  \\
% -      & \checkmark  &  \checkmark             & 6.00  & 7.70   \\
% \checkmark      & \checkmark  &\checkmark            & 5.36 & 6.92 \\          

% % \hdashline
% % \checkmark     &    -          &\checkmark   & \checkmark  & \hm{15.47}  & 99.73  \\ 
% % \checkmark     &    -          &\checkmark   & \checkmark  & 31.03  & 101.89  \\ 
% \bottomrule
% \end{tabular}
% \caption{Performance with different components. SP: superpixel, TPU: transductive prototype updating, MS: multi-scale.}
% \label{tab:aba} 
% \end{table}
% \input{tables/tab_component_carpk}

In this section, we are interested in ablating our approach from the following perspective views on FSC-147 dataset:

% \subsection{Effectiveness of different components}
\noindent \textbf{Effectiveness of different components:} 
% Table~\ref{tab:aba} presents the contribution of each component in our approach. Concretely, comparing the result of the first three rows, both the introduction of superpixel and semantic-rich feature representation (i.e., DINOv2) bring performance gain compared to the SAM baseline (first row). The following two rows respectively demonstrate the contributions of the transductive prototype updating and the multi-scale mechanism to our model performance. Combining all four components together achieves an overall best counting results. Furthermore, the results in the last row show that even without DINOv2, SAM with the proposed other three components can also achieve a decent performance.
% using superpixel-based object-prior object prompts for SAM, the counting MAE decreases from \hm{27.97} to 21.17. 
Table~\ref{tab:aba} showcases the impact of each component in our method in both dataset. Specifically, 
% \hm{as seen in the first three rows, the integration of super-pixel and the semantically rich DINOv2 feature representation both contribute to enhanced performance, surpassing the SAM baseline (first row).} 
although SAM with superpixel can achieve higher interested-object recall (shown in Figure~\ref{fig:superpixel_vs_normal}), due to the lack of semantic differentiation ability in SAM features (shown in Figure~\ref{fig:sam_vs_dino}), simply replacing default point grid with superpixel does not necessarily bring performance gains, as the results shown in the first two rows of Table~\ref{tab:aba}. The next two rows show that using DINOv2 feature representation for SAM's mask proposal selection significantly reduces the final counting error and the combing of superpixel can further bring performance gains.  
The subsequent two rows of Table~\ref{tab:aba} highlight the individual benefits of transductive prototype updating and the multi-scale mechanism on our model's effectiveness. When these four components are combined, they lead to the most impressive counting results, \textcolor{black}{especially in the RMSE metric.}. 
% In contrast to FSC-147, the CARPK dataset proves to be more conducive to our counting task, particularly in scenarios characterized by standard scale and uncluttered scenes. Consequently, the impact of individual components is markedly discernible.
% Additionally, the final row illustrates that 
% incorporating the proposed three components with SAM also yields commendable performance.
% SAM, when combined with the other three proposed components but without DINOv2, also yields commendable performance.

\begin{table}[t]
\centering
\color{black}
\resizebox{0.95\columnwidth}{!}{
\begin{tabular}{c|ccc}
\toprule
number of K & MAE $\downarrow$   & RMSE $\downarrow$  &avg. time/image\\
\hline
% 40*40               & 7.51 & 9.21 &4.16s\\
50*50               & 5.09 & 6.66 &5.26s \\              
60*60               & 4.46 & 5.82 &6.55s \\   
64*64               & 4.39 & 5.70 &7.04s\\        
% 70*70               & 4.36 & 5.67 &7.91s \\               

\bottomrule
\end{tabular}
}
\caption{
% Performance with different number of initialized points and corresponding time consumption on CARPK.
Performance of our method with various K on CARPK.
}
\label{tab:points} 
\vspace{-0.5cm}
\end{table}
% \noindent \textcolor{blue}{\textbf{Different number of initialized points:} Table~\ref{tab:points} shows the performance with different number of initialized points and corresponding time consumption on CARPK~\cite{hsieh2017drone} dataset. With only 50*50 points, our methods surpasses previous SOTA in reasonable time. It also indicates that our method is stable with different number of points. }

\noindent \textcolor{black}{\textbf{Different number of initialized points:} Table~\ref{tab:points} shows that with 50*50 points, our method surpasses the previous SOTAs (see Table~\ref{tab:main}) in reasonable time. 
% While 64*64 increases time by 33\%, it improves MAE by 18\%, demonstrating a balanced trade-off between performance and time consumption.
Increasing to 64*64 points further improves the MAE by 18\%, while the computational time remains within seconds, demonstrating our method's ability to achieve significant performance gains with manageable time costs as the point density increases.
}

\smallskip
\begin{table}[t]
\centering
\resizebox{0.95\columnwidth}{!}{
\begin{tabular}{cc|cc}
\toprule
\multicolumn{1}{c}{} & SAM Backbone & MAE $\downarrow$  & RMSE $\downarrow$ \\ 
\hline
\multirow{2}{*}{TFOC~\cite{shi2023training}} & ViT-B & \multicolumn{1}{c}{19.95} & 132.16 \\
& ViT-H & 30.41 & 138.28 \\
\hline
\multirow{2}{*}{Ours} & ViT-B &  14.43 & 87.54 \\
& ViT-H & 12.26 & 56.33 \\
\bottomrule
\end{tabular}}
\vspace{-0.35cm}
\caption{Performance with different SAM backbone.}
\label{tab:SAM_model} 
% \vspace{-0.4cm}
\end{table}
\noindent \textbf{Backbone size of SAM:} 
% \subsection{Backbone size of SAM}
% Table~\ref{tab:SAM_model} compares the performance of TFOC and ours with various ViT backbones. It shows that our methods can perform better when the backbone scales from base to huge, while the TFOC even performs worse when ViT-H is used which may be because more mask proposals are generated with ViT-H but poor proposal selection worsens the performance. Thanks to the use of semantic-rich features, our method benefits from the more generated object proposals. 
Table~\ref{tab:SAM_model} compares TFOC and our method across different ViT backbones, revealing that our approach improves as the backbone scales from base to huge. In contrast, TFOC's performance declines with ViT-H, possibly due to an increase in mask proposals leading to poorer selection. Our method, leveraging semantically rich features, benefits from the higher number of generated object proposals.

% \textbf{Robustness of our approach}\\
% hm: robustness does not necessary --> SAM backbone size
% In practical, we find that for some methods like \cite{shi2023training}, different SAM model caused huge fluctuation (more than 50\% MAE increase) in performance (shown in Table~\ref{tab:SAM_model}).  Here we apply our methods on SAM with different sizes. Table~\ref{tab:SAM_model} indicates that our method is robust in different SAM models. For huge SAM, This results in a 4.6\% drop of MAE and a 29\% increase of RMSE, which indicates that SAM base could generate more masks for multi-scale images. After visualizing the results, the masks from SAM base are not correct, the semantic features of those wrong masks have high similarity as the size of those masks are extremely small (small than 5 pixels).

% \smallskip
% \subsection{Various semantic-rich vision models}
\noindent \textbf{Various semantic-rich vision models:}
% Table~\ref{tab:semantic_model} shows the performance of our method with various vision models and The results show that a stronger vision model can obtain better counting accuracy.
Table~\ref{tab:semantic_model} displays our method's performance across different \textcolor{black}{SOTA foundation vision models}, indicating that superior vision models yield improved counting accuracy. \textcolor{black}{For CLIP, only the vision encoder is involved.}

\begin{table}[t]
\centering
\resizebox{0.95\columnwidth}{!}{
\begin{tabular}{l|ccc}
\toprule
Vision Model & Patch Grid & MAE $\downarrow$   & RMSE $\downarrow$ \\
 \hline
CLIP~\cite{radford2021learning} & 16*16  & \hm{17.83} & \hm{53.81}       \\
DINO~\cite{caron2021emerging} & 28*28 & 14.28 & 59.33 \\
DINO v2~\cite{oquab2023dinov2} & 37*37 & 12.26 & 56.33 \\
\bottomrule
\end{tabular}}
\vspace{-0.35cm}
\caption{Performance with different semantic-rich model.}
\label{tab:semantic_model} 
% \vspace{-0.2cm}
\end{table}
\begin{table}[t]
\centering
\resizebox{0.65\columnwidth}{!}{
\begin{tabular}{c|cc}
\toprule
Update Round & MAE $\downarrow$   & RMSE $\downarrow$  \\
\hline
0               & 13.77 & 87.40 \\
1               & 12.26 & 56.33 \\
2               & 12.30 & 57.50 \\
\bottomrule
\end{tabular}
}
\vspace{-0.35cm}
\caption{Performance with various rounds of TPU.}
\label{tab:no_of_pl} 
\vspace{-0.5cm}
\end{table}

% \input{figures/fig_theta_delta}

% \smallskip
\noindent \textbf{Multi-round transductive prototype updating:}
% \subsection{Multi-round transductive prototype updating}
% In our work, one round of transductive prototype updating is conducted by default. Results in Table~\ref{tab:no_of_pl} show that more round of updating does not bring performance gain.
Our study typically employs one round of transductive prototype updating (TPU). Table~\ref{tab:no_of_pl} demonstrates that additional rounds do not enhance performance.

% \smallskip
% \noindent \textbf{Threshold sensitivity:}
% % \subsection{Threshold sensitivity}
% % In our work, there are two hyperparameters in our approach: similarity threshold $\theta$ for proposal selection and $\delta$ for transductive prototype updating. Table~\ref{} shows that our approach is robust to the selection of both hyperparameters.
% Our method involves two hyperparameters: a similarity threshold $\theta$ for proposal selection and $\delta$ for transductive prototype updating. Figure~\ref{fig:theta_delta} illustrates the robustness of our method to variations in the hyperparameters.

\section{Conclusion}
\label{sec:Conclusion}
% In conclusion, our comprehensive study presents a novel framework that significantly enhances Class-Agnostic Counting (CAC) without the dependence on extensive training data. By integrating superpixel-based precision in anchor point placement, semantically enriched image encoders, a multiscale mechanism for minute object counting, and a transductive prototype update strategy, we have established a robust, training-free CAC methodology. Our empirical results on the large-scale dataset FSC147 and car counting dataset CARPK underscore the efficacy of this approach. Not only achieve state-of-the-art performance, we also bridge the performance gap between training-free and training-based methods.
This paper presents a novel, training-free approach for Class-Agnostic Counting (CAC), significantly closing the performance gap with trained models. Key contributions include the use of superpixels, a richer semantic image encoder, a multiscale mechanism, and a transductive prototype updating strategy. 
% Experiments validate these innovations, showcasing that training-free CAC can match the performance of trained approaches. This work marks a pivotal advancement in CAC methodologies, offering a new benchmark and highlighting the potential of training-free methods in computer vision.
% Experimental results affirm these innovations, proving training-free CAC can equal trained model performance. This research sets a new standard in CAC, demonstrating the untapped potential of non-training approaches. 
% Additionally, it paves the way for future advancements in machine learning, especially in areas where training data is scarce or unavailable. This approach could revolutionize the way we approach object counting and analysis in computer vision.
Through rigorous experimental evaluation, the effectiveness of these four key components are demonstrated. The research shows that a CAC model without any training can achieve performance on par with training-based CAC approaches. This is a significant advancement in the field, indicating the potential of training-free approaches and setting a new benchmark for future research in CAC.

\newpage
%%%%%%%%% REFERENCES
{\small
\bibliographystyle{ieee_fullname}
\bibliography{egbib}
}

\end{document}

%%%%%%%%% TITLE - PLEASE UPDATE
\title{Supplementary Material for A Simple-but-effective Baseline for Training-free Class-Agnostic Counting}

\author{Yuhao Lin\footnotemark[1] , Haiming Xu\thanks{Equally contributed.} , Lingqiao Liu, Javen Qinfeng Shi\\
Australian Institution for Machine Learning\\
The University of Adelaide\\
{\tt\small yuhao.lin01,hai-ming.xu,lingqiao.liu,javen.shi@adelaide.edu.au}
% For a paper whose authors are all at the same institution,
% omit the following lines up until the closing ``}''.
% Additional authors and addresses can be added with ``\and'',
% just like the second author.
% To save space, use either the email address or home page, not both
% \and
% Second Author\\
% Institution2\\
% First line of institution2 address\\
% {\tt\small secondauthor@i2.org}
}
\maketitle

In this supplementary material, we provide more details to complement the manuscript, including threshold sensitivity, visualization of generated mask proposal and qualitative results on CARPK dataset.

% \vspace{-0.2cm}
\section{Threshold Sensitivity}

\begin{figure}[h]
\centering
\vspace{-0.2cm}

\includegraphics[width=0.48\linewidth]{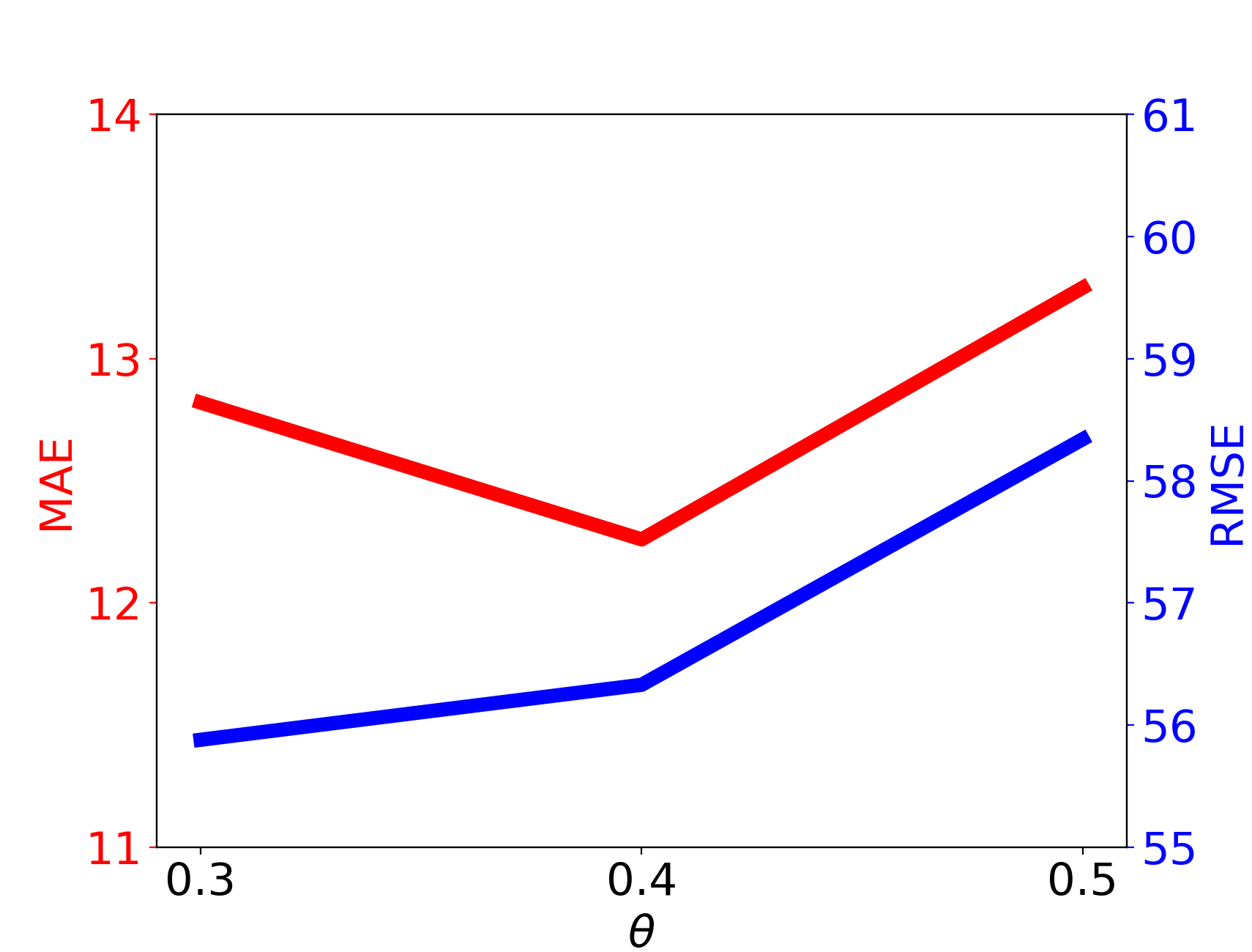}  
\includegraphics[width=0.48\linewidth]{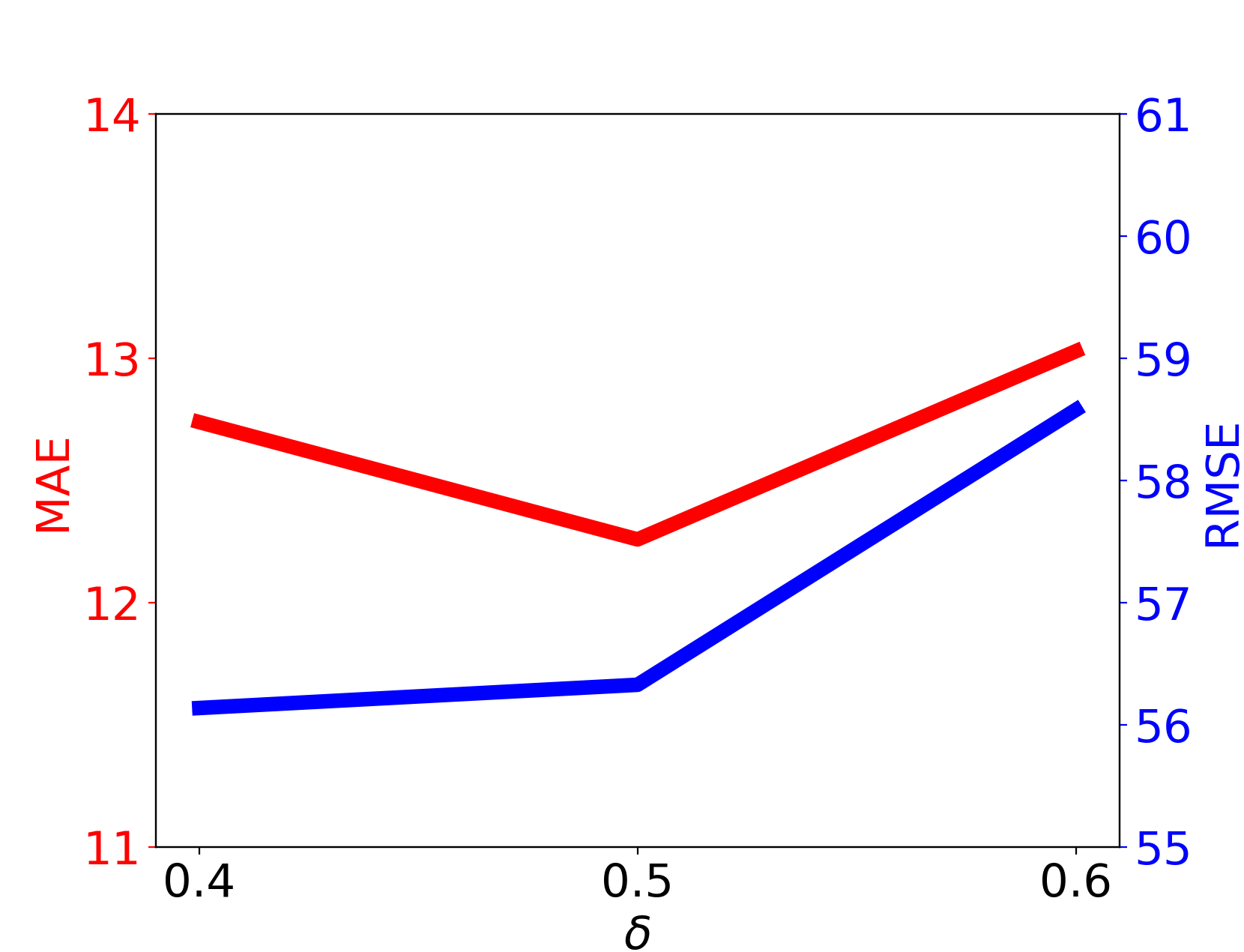}  

\caption{Performance with various threshold $\theta$ and $\delta$.}
\label{fig:theta_delta}
\vspace{-0.3cm}
\end{figure}
% \subsection{Threshold sensitivity}
% In our work, there are two hyperparameters in our approach: similarity threshold $\theta$ for proposal selection and $\delta$ for transductive prototype updating. Table~\ref{} shows that our approach is robust to the selection of both hyperparameters.
\textcolor{black}{Our method involves two hyperparameters: a similarity threshold $\theta$ for proposal selection and $\delta$ for transductive prototype updating. Figure~\ref{fig:theta_delta} illustrates the robustness of our method to variations in the hyperparameters.}

% \vspace{-0.2cm}

% \section{Various Superpixel Methods}
% \textcolor{black}{Superpixels were chosen for
% their ability to group pixels into meaningful regions aligning with object boundaries, making them suitable for generating object-prior point prompts for SAM. Our comparative
% analysis (Tab. ~\ref{tab:sp_supp}) showed SLIC superpixels offer an optimal balance of performance and efficiency. We used 4096
% clusters (64x64) as the SLIC parameter, providing sufficient
% granularity while maintaining low runtime.}
% \begin{table}[h]\tiny
% \centering
% \color{black}
% \resizebox{0.95\columnwidth}{!}{
% \begin{tabular}{c|ccc}
% % \toprule
% Superpixel & MAE $\downarrow$   & RMSE $\downarrow$  & avg. time/image\\
% \hline
% SLIC               & 12.25 & 56.36 & 10 ms  \\
% SNIC               & 14.26 & 65.20 & 3.68 s \\
% % \bottomrule
% \end{tabular}
% }

% \caption{Our method with various superpixels on FSC-147.}
% \label{tab:sp_supp} 
% \vspace{-0.4cm}

% \end{table}
% \vspace{-0.2cm}

\section{Comparison with PseCo}
\begin{figure}[t]
\centering

\includegraphics[width=0.95\linewidth]{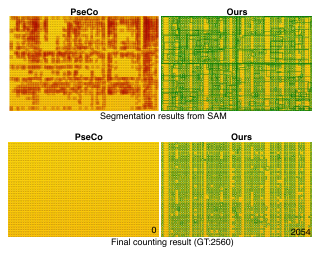}

\caption{FSC-147 example results. Zoom in for detail.}
\label{fig:vsPseCosupp}
% \vspace{-0.6cm}
\end{figure}

\textcolor{black}{
Both methods leverage foundation models for CAC, using SAM for object segmentation and semantic-rich encoders for final counting decisions. Key differences include:} \textcolor{black}{\textbf{Approach:} 
Our method demonstrates foundation model capabilities without additional trained modules, while PseCo incorporates two extra trained components (point decoder and object classifier).} \textcolor{black} {\textbf{Performance:}Our method outperforms PseCo on FSC-147, especially in RMSE (56.33 vs 112.86). This improvement is notable in images with crowded objects. As shown in Fig.~\ref{fig:vsPseCosupp}, when counting Lego block protrusions (2560 total), PseCo's inaccurate point decoder leads to poor SAM segmentation and its object classifier predicts a count of 0. In contrast, our method, using superpixels for object-prior point prompts and a prototype-based classifier, identifies 2054 objects. }

% \vspace{-0.2cm}

\section{Visualization of Generated Mask Proposal}

\begin{figure*}[h]
\centering
\includegraphics[width=\linewidth]{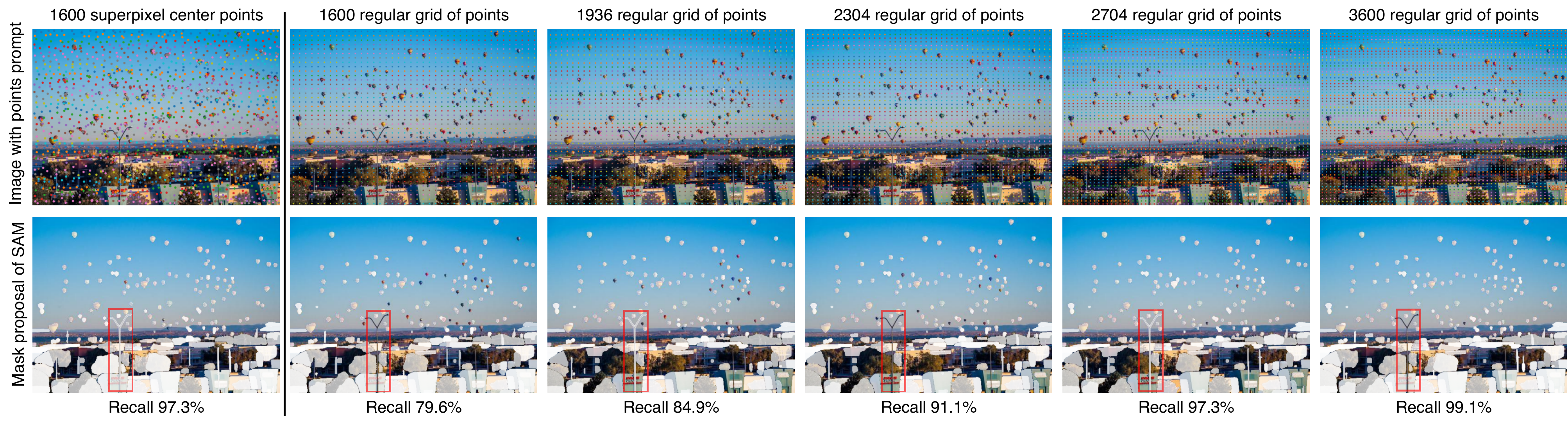}
\caption{Visualization of mask proposals generated by SAM with various point prompts. In this example, the hot-air balloon is the user-interested object and the bottom row presents the recall of this object. The red boxes display mask generation for a very fine object (the street lamp).
% It clearly shows that SAM with superpixel can achieve a high recall rate while maintaining a lower time cost compared to the default point grid prompts.  utilizing superpixel for SAM significantly enhances the interested object recall rate without adding to the quantity of point prompts, effectively bypassing any increase in computational load.
}
\label{fig:masks}
\end{figure*}

In the main paper, we have presented the effectiveness of the use of superpixel on the quality of mask proposals generated by SAM in Figure 3. In this section, we further illustrate the visualization of generated mask proposals in Figure~\ref{fig:masks}. 
% the mask proposal and its corresponding recall rate from different numbers of points ranging from 1600 to 3600. 
It clearly shows that SAM with superpixel can achieve a high recall rate for the interested object without demanding of denser grid of points. Meanwhile, we find that SAM with superpixel can also generate mask proposals for very thin objects thanks to the object-prior prompt, such as the street lamp marked in the red box. Nonetheless, employing SAM with a standard grid of points might not consistently achieve segmentation for such objects.

\section{Qualitative Results on CARPK}\label{sec:vis-carpk}
In this section, we further provide some qualitative visualization for the CARPK dataset. As shown in Figure \ref{fig:car}, our method can generate precise masks for each car instance and achieve accurate counting results.

\begin{figure*}[h]
\centering
\includegraphics[width=0.98\linewidth]{images/carpk_mm.pdf}  
\caption{
Qualitative results on the CARPK dataset are displayed. Counting values are noted in the bottom-left corner. 
% Our method accurately predicts masks in both densely and sparsely populated scenes. 
Best viewed by zooming in.}
\label{fig:car}
\end{figure*}
%%%%%%%%% REFERENCES
% {\small
% \bibliographystyle{ieee_fullname}
% \bibliography{egbib}
% }